\documentclass[letterpaper]{article} 
\usepackage{aaai2027}  
\usepackage[hyphens]{url}  
\usepackage{graphicx} 
\usepackage{natbib}  
\usepackage{caption} 
\usepackage{algorithm}
\usepackage{algorithmic}

\usepackage{pgfplots}
\usepackage{subcaption}
\pgfplotsset{compat=1.17}

\definecolor{myorange}{RGB}{233, 114, 77}  
\definecolor{myteal}{RGB}{42, 157, 143}    
\definecolor{mydark}{RGB}{38, 70, 83}      
\usepackage{amssymb}
\usepackage{amsmath}
\usepackage{pifont}
\newcommand{\cmark}{\ding{51}}

\usepackage{newfloat}
\usepackage{listings}
\DeclareCaptionStyle{ruled}{labelfont=normalfont,labelsep=colon,strut=off} 
\floatstyle{ruled}
\newfloat{listing}{tb}{lst}{}
\floatname{listing}{Listing}

\usepackage{booktabs}

\title{When Diffusion Models Forget Who You Are: Identity Preservation in Face Inpainting under Large Occlusions}
\author{Feng Ding, Shuhuai Xie, Yue Zhou, Yulan Zhang, Guopu Zhu, Mengyao Xiao}
\affiliations{Nanchang University, Shenzhen University, Huizhong University, Harbin Institute of Technology}

\begin{document}

\maketitle

\begin{abstract}
Face inpainting with diffusion models has recently achieved impressive visual quality, yet preserving identity fidelity under significant occlusion and conflicting text guidance remains a major challenge. To address this issue, we present Reference Semantic Inpainting for Face (ReSem-Face), a cascaded diffusion framework that introduces an explicit identity-conditioned semantic prior for multi-reference face inpainting. Our approach distills representative identity features from multiple references to reconstruct missing semantic regions, which then guide the diffusion process through a multi-stream conditioning architecture. This design provides strong semantic constraints when pixels are absent and stabilizes identity reconstruction while remaining compatible with prompt-driven edits. Experiments on CelebAHQ-IDI-5 and VGGFace2 demonstrate that ReSem-Face yields more reliable identity-preserving completion under severe semantic masks and improves text-controlled editing quality compared with representative baselines.
\end{abstract}

\section{Introduction}

Face inpainting \cite{ju2024brushnet,zhuang2024task,xiao2025omnigen} refers to the task of completing missing or occluded facial regions while preserving the identity of a specific person and maintaining natural consistency with the visible context (e.g., expression, and background). Unlike generic inpainting, the goal is not only to produce a plausible face but to reconstruct the same face: subtle geometry and fine-grained textures that make a person recognizable must be retained even when key identity cues are severely removed by the mask.

Early works typically build on convolutional or GAN-based inpainting pipelines\cite{cai2019fcsr,li2017generative,goodfellow2014generative}, sometimes augmented with facial priors such as landmarks, parsing maps, or multi-scale discriminators to enhance realism and structural plausibility. These methods achieve visually coherent completions under mild to moderate occlusions and often produce sharp textures. However, when the missing region is large or covers identity-critical areas (e.g., eyes, nose, and mouth), they tend to hallucinate person-agnostic details, leading to identity drift; moreover, their reliance on local context makes them brittle under expression variations and hard to extend to controllable edits.

It is worth noting that diffusion models\cite{zhang2023adding,wu2023tune,zhao2023uni,zhu2024sd,wasserman2025paint,xie2025turbofill} have recently advanced image synthesis and editing by progressively denoising from noise to data, offering strong generative priors and impressive fidelity in texture and global consistency. Motivated by this, diffusion-based inpainting frameworks \cite{avrahami2022blended,avrahami2023blended,zhang2023adding,kim2025rad,huang2025mtadiffusion} have emerged as a promising direction for face restoration, and reference-guided variants further attempt to inject identity information from one or more reference images \cite{luo2023reference,xu2024personalized}so that the denoising process can reconstruct the correct subject rather than an average face.

Despite the success of reference-guided diffusion and personalized generation techniques, robust identity preservation under heavy occlusion remains challenging. In practice, reference images are often misaligned with the target, as the masked target  offers little direct evidence for identity. More importantly, many existing designs inject identity cues implicitly during denoising and expect them to simultaneously determine who the subject is and what content should fill the missing region\cite{ruiz2023dreambooth,li2024photomaker,gal2023an,kumari2023multi}; under high uncertainty and noise, this coupling can weaken identity signals, and the problem becomes even harder when text-guided editing \cite{zhang2020text1,nichol2021glide,zhang2020text,xie2023smartbrush,chen2024improving} is introduced, where semantic instructions may inadvertently override or distort identity-specific details.

Motivated by the aforementioned limitations, we propose Reference Semantic Inpainting for Face (ReSem-Face), a diffusion-based~\cite{rombach2022high} personalized face inpainting framework that introduces an explicit identity-conditioned semantic prior to stabilize identity reconstruction under severe occlusion. Specifically, we design an Identity-Aware Semantic Pre-Inpainting module that aggregates cues from multiple references in a clean feature space and predicts semantic tokens describing the missing region’s identity-relevant structure and appearance; these tokens are then injected into the diffusion U-Net via dedicated semantic attention, operating alongside text guidance and an introduced Reference Identity Attention pathway to form complementary constraints. By decoupling identity-related semantics from the noisy denoising trajectory, ReSem-Face enables more faithful identity preservation and more reliable controllable editing, especially in challenging large-mask scenarios.

In summary, our contributions are as follows: 
\begin{itemize}
    \item We propose a novel Identity-Aware Semantic Pre-Inpainting Module that predicts identity-aware geometry and texture semantics for missing facial regions, forming a high-level semantic prior for diffusion-based inpainting.
    \item We integrate this semantic prior into the diffusion U-Net through a Reference Semantic Attention pathway, which works together with text cross-attention and Reference Identity Attention to enable tri-conditioning with text, identity, and semantic guidance.
    \item Extensive experiments demonstrate that ReSem-Face achieves state-of-the-art identity preservation and strong controllability in both standalone face inpainting and text-guided editing scenarios.
    
\end{itemize}

\section{Related Work}
\subsection{Face Inpainting}
The past five years have witnessed rapid progress in face inpainting, driven by the demand to realistically recover occluded facial regions while keeping identity and fine-grained attributes \cite{wang2021towards,xu2024personalized,suvorov2022resolution}. Early works focus on identity-guided completion under heterogeneous domains, typically combining staged fitting and refinement to better align the synthesized face with the surrounding context   \cite{li2021faceinpainter,luo2023reference}. To this end,  Mask-aware transformer (MAT) \cite{li2022mat,vaswani2017attention} utilizes transformer-style long-range modeling to fill large holes with coherent global structure at high resolution. Due to the development of deep generative models, diffusion-based priors have been adopted to improve semantic plausibility and output diversity without retraining for specific mask types \cite{lugmayr2022repaint,chen2024improving,luo2023reference,rombach2022high}. Some works \cite{dong2022incremental,luo2023reference} propose to explicitly restore structural cues (e.g., sketches and edges) and inject them into texture completion, strengthening geometry–texture consistency. Furthermore, \cite{suvorov2022resolution} proposes to use Fourier-based global receptive fields to better handle large irregular masks and high-resolution images. However, despite these advances, person-specific fidelity can still drift—subtle identity cues may change when the missing region is large—motivating more identity-aware face inpainting designs \cite{motamed2023patmat}.

\subsection{Reference-guided Face Inpainting}
A widely adopted and highly effective solution to improve identity fidelity in face inpainting is to condition the model on one or multiple reference images of the same subject\cite{zhou2021transfill,luo2023reference,motamed2023patmat,varanka2024pfstorer}, which provide identity-critical cues that are absent in heavily masked targets. In early representative efforts, reference-attention–based designs \cite{yu2022reference} explicitly align and fuse reference features with the corrupted input, encouraging the completion to resemble the reference identity more faithfully. Following that, dual-control formulations further disentangle the reference signal into high-level identity and low-level texture \cite{luo2023reference}, enabling more controllable and higher-quality completion under large-scale missing regions. PVA (Xu et al. 2024) integrates Parallel Visual Attention into a pretrained diffusion inpainting model, injecting reference-image features into the denoising network to achieve identity-preserving and language-controllable face inpainting with lightweight per-identity tuning. Recent works (Yang et al. 2023; Chen et al. 2024a; Motamed et al. 2023) also conduct reference guidance via personalization-by-tuning and diffusion-based exemplar conditioning for stronger reference-driven generation and editing; meanwhile, reference-guided methods have also been extended to directional and diverse face inpainting to produce multiple plausible yet reference-consistent outcomes. Nevertheless, most existing reference-guided methods mainly introduce identity cues as denoising-time conditions, while the semantic content of heavily occluded regions is still inferred implicitly during diffusion. In contrast, our ReSem-Face explicitly predicts identity-aware semantic priors before diffusion denoising, providing additional geometry and texture constraints for large-mask identity reconstruction.

\section{Methods}
We begin by noting that, in multi-reference identity-preserving face inpainting, a plausible synthesis needs to be generated for a large missing facial region while remaining faithful to the identity of a specific person provided by a set of reference images. This setting becomes even more challenging when the completion is further guided by a text prompt for attribute editing, since identity consistency and prompt compliance can be conflicting under severe occlusions. In this section, we will delve into the details of our proposed ReSem-Face framework, and the pipeline of ReSem-Face is shown in Fig.~\ref{fig:framework}. We first provide a brief review of latent diffusion inpainting, and then present the identity-aware semantic pre-inpainting module, the Reference Identity Attention pathway, and the ReSemAttn-based injection scheme for semantic priors, followed by the training objectives and implementation details.

\subsection{Preliminaries}
Diffusion-based generative models view image synthesis as learning to reverse a gradual noising process.  
Starting from a clean sample $x_0$, the forward process constructs a sequence $\{x_t\}_{t=1}^T$ by repeatedly injecting Gaussian noise with a variance schedule $\{\beta_t\}_{t=1}^T$:
\begin{equation}
x_{t+1} = \sqrt{1-\beta_t}\,x_t + \sqrt{\beta_t}\,\epsilon,\qquad \epsilon \sim \mathcal{N}(0, I),
\end{equation}
where $0 < \beta_t < 1$. After sufficiently many steps, $x_t$ becomes nearly Gaussian and the original data is largely destroyed.  
By composing the forward transitions, one can directly relate $x_t$ to $x_0$ in closed form:
\begin{equation}
x_t = \sqrt{\bar{\alpha}_t}\,x_0 + \sqrt{1-\bar{\alpha}_t}\,\epsilon_t,
\end{equation}
with $\bar{\alpha}_t = \prod_{i=1}^t (1-\beta_i)$ denoting the cumulative signal decay.  
The generative model is then defined by a denoising network $\epsilon_\theta(x_t, t)$ that predicts the noise added at each step.  
It is trained using a denoising score-matching objective
\begin{equation}
\mathcal{L}_{\mathrm{DSM}} = \mathbb{E}_{x_0, t, \epsilon \sim \mathcal{N}(0, I)} 
\big[\|\epsilon - \epsilon_\theta(x_t, t)\|_2^2\big],
\end{equation}
where $t$ is sampled uniformly from $\{1,\dots,T\}$ and $x_t$ is obtained from $x_0$ via the forward process.  
At inference time, the model approximately inverts the diffusion chain, starting from Gaussian noise and iteratively denoising it.  
In practice, we adopt a DDIM-style sampler for efficient generation\cite{song2020denoising}.

Running this procedure directly in pixel space is computationally demanding.  
Latent Diffusion Models (LDMs)\cite{rombach2022high} address this limitation by applying the same diffusion formulation in a compressed latent representation.  
Let $E_V(\cdot)$ and $D_V(\cdot)$ denote the encoder and decoder of a Variational Auto-Encoder (VAE) \cite{kingma2013auto}, and $z_t = E_V(x_t)$, $x_t = D_V(z_t)$ be the encoding and decoding operations.  
The diffusion model is then defined over latent variables $z_t$ instead of raw images.  
Moreover, LDMs are typically conditioned on text through cross-attention layers that attend to language features $y_i = E_T(T_i)$ extracted by a pretrained text encoder (e.g., CLIP\cite{radford2021learning}).  
For inpainting, the latent model also receives the occluded image and the corresponding binary mask as additional inputs.  
Concretely, we form a mask-aware latent input
\begin{equation}
\tilde{z}_t = z_t \,\|\, u_{\downarrow}(m) \,\|\, E_V(m \odot x_0),
\end{equation}
where $m$ is the mask, $u_{\downarrow}(m)$ denotes downsampling $m$ to the latent resolution, $\odot$ is element-wise multiplication, and $\|\,$ indicates channel-wise concatenation.  
The inpainting LDM is then trained with a noise-prediction loss analogous to the DDPM objective:
\begin{equation}
\mathcal{L}_{\mathrm{LDM}} = 
\mathbb{E}_{z_0, y, m, t, \epsilon}
\big[\|\epsilon - \epsilon_\theta(\tilde{z}_t, y, t)\|_2^2\big],
\end{equation}
where $z_0 = E_V(x_0)$ and $y$ denotes the text conditioning.  
Our method builds on top of such a latent diffusion inpainting backbone, which jointly exploits the text prompt, the masked image, and the mask to guide the denoising process.

\subsection{Identity-Aware Semantic Pre-Inpainting Module}
\begin{figure*}[t]
  \centering
  \includegraphics[trim=0 0 10pt 0, clip, width=\textwidth]{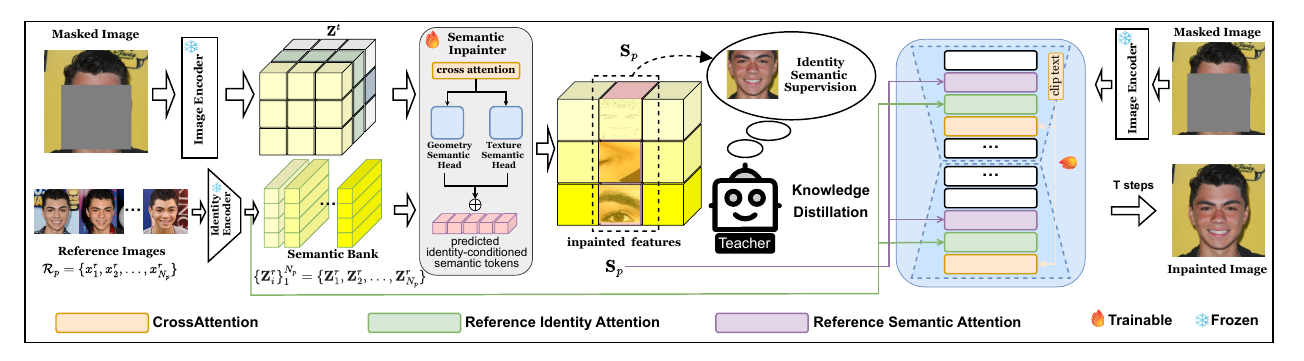}
  \caption{
    The pipeline consists of three main components: 
    (1) Identity Semantic Bank Construction, which aggregates identity features from multiple reference images; 
    (2) Semantic Pre-Inpainting, where a semantic inpainter predicts semantic tokens for the masked region in the clean feature space; and 
    (3) Tri-Conditioned Diffusion, where the predicted tokens are injected into the denoising backbone via the Reference Semantic Attention (ReSemAttn) pathway, which operates in parallel with text cross-attention and Reference Identity Attention to guide the generation.
  }
  \label{fig:framework}
\end{figure*}

To provide the diffusion backbone with high-level priors describing the latent structure of the occluded facial region, we introduce an Identity-Aware Semantic Pre-Inpainting Module. This module forms an additional semantic pathway complementary to Reference Identity Attention, producing identity-conditioned semantic tokens that encode both geometry and fine-grained appearance.  
The overall design follows a cascaded \cite{vaswani2017attention} architecture reminiscent of CAT-Diffusion but differs fundamentally in its purpose, semantic targets, and integration strategy.
The full framework of ReSem-Face is shown in Fig.~\ref{fig:framework}

\subsubsection{Multi-Reference Identity Semantic Aggregation}
Given a set of reference images $\mathcal{R}_p=\{x_i^r\}_{i=1}^{N_p}$ from the same identity, we first extract per-reference identity-aware tokens using a frozen identity encoder $E_{\mathrm{id}}$, where the FaceNet branch is implemented as a ResNet-50 model trained with ArcFace~\cite{deng2019arcface}. To suppress intra-identity variations, we perform cross-image token interaction via multi-head attention over all reference tokens, aggregating them into a unified identity semantic bank:
\begin{equation}
\mathbf{Z}^{r}=\mathrm{Norm}\!\left(\frac{1}{N_p}\sum_{i=1}^{N_p}\mathrm{MSA}\!\big(\mathbf{Z}^{r}_i,\mathbf{Z}^{r}_{1:N_p},\mathbf{Z}^{r}_{1:N_p}\big)\right),
\end{equation}
where $\mathbf{Z}^{r}_i=E_{\mathrm{id}}(x_i^r)$, and $\mathbf{Z}^r$ serves as a consistent identity memory queried by the masked target in subsequent stages.

\subsubsection{Identity-conditioned semantic pre-inpainting}
Given a masked target image $x^t$ and its binary mask $m$, we encode the visible context with a CLIP-based mask-aware visual encoder $E_{\mathrm{vis}}$ to obtain target tokens $\mathbf{Z}^t$. We then fuse $\mathbf{Z}^t$ with the identity bank $\mathbf{Z}^r$ through multi-head cross-attention, so that identity cues can be propagated into partially observed or fully missing regions. The fused representation is decoded by a lightweight transformer decoder with two heads: (i) a geometry head that predicts structure-oriented semantics (e.g., component layout and shape consistency), and (ii) a texture head that predicts identity-dependent appearance traits. Their outputs are finally projected into a fixed-length sequence of identity-conditioned semantic tokens $\mathbf{S}_p$ for the occluded region.

\subsubsection{Identity-aware semantic supervision}
To ensure $\mathbf{S}_p$ captures both semantic correctness and identity consistency, we supervise the geometry/textural predictions and enforce identity alignment using features extracted from the ground-truth region $x^{\mathrm{gt}}$. In addition, we adopt a frozen teacher model to provide stronger semantic targets and distill high-level knowledge into $\mathbf{S}_p$, which stabilizes semantic prediction under large occlusions and improves prompt alignment:
\begin{equation}
\begin{aligned}
\mathcal{L}_{\mathrm{pre}}
&=\lambda_{\mathrm{id}}\!\left(1-\cos\!\left(\phi_{\mathrm{id}}(x^{\mathrm{gt}}),\,\phi_{\mathrm{id}}(\mathbf{S}_p)\right)\right)
+\lambda_{\mathrm{geo}}\left\|\phi_{\mathrm{geo}}(x^{\mathrm{gt}})-\mathbf{G}\right\|_{1} \\
&\quad+\lambda_{\mathrm{tex}}\left\|\phi_{\mathrm{tex}}(x^{\mathrm{gt}})-\mathbf{T}\right\|_{1}
+\lambda_{\mathrm{t}}\left\|\psi_{\mathrm{t}}(x^{\mathrm{gt}})-\psi_{\mathrm{t}}(\mathbf{S}_p)\right\|_{1}.
\end{aligned}
\end{equation}

where $\phi_{\mathrm{id}},\phi_{\mathrm{geo}},\phi_{\mathrm{tex}}$ extract identity/geometry/texture descriptors, $\mathbf{G}$ and $\mathbf{T}$ denote the geometry/texture head outputs, and $\psi_{\mathrm{t}}(\cdot)$ denotes teacher features used for distillation\cite{navaneet2022simreg}.

\subsubsection{Reference Identity Attention}
To complement the semantic prior with direct identity evidence, we introduce a Reference Identity Attention (RIA) pathway for identity-conditioned denoising. Given the aggregated reference identity tokens $\mathbf{H}_p$, RIA lets the intermediate U-Net features selectively retrieve identity-specific cues from the reference set and injects them into the denoising process as an identity guidance branch. Different from the semantic pathway, which predicts explicit geometry and texture priors for the missing region, RIA focuses on preserving subject-level identity consistency during generation. In this way, identity retrieval and semantic prior injection are modeled as two complementary conditions within the tri-conditioning framework.

\subsubsection{Reference Semantic Attention for Diffusion Conditioning}
We inject the predicted semantic tokens $\mathbf{S}_p$ into the diffusion U-Net via a dedicated Reference Semantic Attention (ReSemAttn) pathway. At each transformer block, ReSemAttn operates in parallel with text cross-attention and Reference Identity Attention (RIA), providing complementary constraints to guide denoising:
\begin{equation}
\begin{split}
\mathbf{Z}_{\ell+1}
&= \mathrm{SelfAttn}(\mathbf{Z}_{\ell})
+\mathrm{CrossAttn}_{\mathrm{text}}(\mathbf{Z}_{\ell},\mathbf{Y}) \\
&\quad + \mathrm{RIA}(\mathbf{Z}_{\ell},\mathbf{H}_p)
+\mathrm{ReSemAttn}(\mathbf{Z}_{\ell},\mathbf{S}_p),
\end{split}
\end{equation}

where $\mathbf{Z}_{\ell}$ is the hidden state at block $\ell$, $\mathbf{Y}$ is the text condition, and $\mathbf{H}_p$ denotes the reference identity tokens. This tri-pathway conditioning encourages identity-faithful and semantically coherent completion, especially when visible evidence is sparse.

\subsection{Training}
Following the standard latent diffusion inpainting formulation, ReSem-Face optimizes a diffusion-based inpainting objective augmented with reference identity conditioning and identity-aware semantic supervision.

\subsubsection{Diffusion Loss}
We adopt the standard latent diffusion objective, where the model predicts noise added at timestep $t$. Given masked latent $\tilde{\mathbf{z}}_t$, text prompt $\mathbf{y}$, reference identity tokens $\mathbf{H}_p$, and semantic tokens $\mathbf{S}$, the denoising loss is:
\begin{equation}
\mathcal{L}_{\mathrm{diff}} =
\mathbb{E}_{\mathbf{z}_0, t, \epsilon}
\left[
\left\|
\epsilon -
\epsilon_\theta(\tilde{\mathbf{z}}_t, t, \mathbf{y}, \mathbf{H}_p, \mathbf{S})
\right\|_2^2
\right].
\end{equation}
This objective aligns all three conditioning streams (text, identity, semantic) with the diffusion trajectory.

\subsubsection{Identity Preservation Loss}
To ensure that personalized details are preserved in the generated face, we introduce an \emph{identity-consistency loss} between the reconstructed image $\hat{x}$ and the ground-truth $x^{\mathrm{gt}}$. Using a pretrained identity extractor $\phi_{\mathrm{id}}$, we compute:
\begin{equation}
\mathcal{L}_{\mathrm{id}} =
1 - \cos\!\left(
\phi_{\mathrm{id}}(\hat{x}),
\phi_{\mathrm{id}}(x^{\mathrm{gt}})
\right).
\end{equation}
This constraint guides the diffusion model toward identity-faithful synthesis, especially in large-missing-region cases.

\subsubsection{Reference Semantic Loss}
Our Identity-Aware Semantic Pre-Inpainting Module predicts geometry and texture semantic tokens $(G, T)$ for the occluded region.  
To align predicted semantics with the ground truth region’s structure and appearance, we propose a \emph{Reference Semantic Loss}:
\begin{equation}
\mathcal{L}_{\mathrm{sem}} =
\lambda_1 \left\|\phi_{\mathrm{geo}}(x^{\mathrm{gt}}) - G\right\|_1
+ \lambda_2 \left\|\phi_{\mathrm{tex}}(x^{\mathrm{gt}}) - T\right\|_1.
\end{equation}
These terms supervise the semantic branch to produce accurate high-level priors before diffusion denoising.

\section{Experiments}

\subsection{Experiment Settings}
\subsubsection{Dataset and Pre-processing}
We conduct experiments on CelebAHQ-IDI-5~\cite{xu2024personalized}, a tailored benchmark for multi-reference face inpainting containing 1,963 identities, and VGGFace2~\cite{cao2018vggface2}, a large-scale dataset comprising over 3.3 million images across 9,131 identities. Unlike the aligned CelebA-HQ, VGGFace2 serving as a challenging testbed to evaluate our model's generalization capability in unconstrained "in-the-wild" scenarios.
Following standard protocols, we utilize 5 reference images per identity and evaluate on unseen identities. 
To mimic real-world occlusions, we apply semantic masks including lower-face, eye\&brow, whole-face, and random regions. 
The input consists of a masked target image, the corresponding binary mask, and the reference set. 
All images are kept with original alignment and resized to the diffusion backbone's input resolution. 
During training, we randomly sample masks to enhance robustness, while evaluation is performed on specific mask categories to assess performance under diverse occlusion semantics.

\subsubsection{Implementation Details}
ReSem-Face follows a two-stage training scheme. 
Stage I optimizes the semantic pre-inpainter via $\mathcal{L}_{\mathrm{pre}}$ using Adam with a learning rate of $1\times 10^{-5}$ for 20K iterations on 4 A40 GPUs. 
Stage II finetunes the diffusion backbone. 
We freeze the original U-Net layers and the Identity Encoder, and train only the introduced modules, including ReSemAttn, and Reference Identity Attention, using the joint objective of $\mathcal{L}_{\mathrm{diff}}$, $\mathcal{L}_{\mathrm{id}}$, and $\mathcal{L}_{\mathrm{sem}}$. 
This stage runs for 200K iterations using AdamW with a learning rate of $1.6\times 10^{-5}$ and a weight decay of $10^{-2}$, employing classifier-free guidance by dropping conditioning with a probability of 0.1. 
At inference, we perform a lightweight personalization of 40 steps for each target identity.

\subsubsection{Evaluation Metrics}
We compare ReSem-Face against eight baselines including LDI~\cite{rombach2022high}, Custom Diffusion~\cite{kumari2023multi}, Textual Inversion~\cite{gal2023an}, ReF-LDM~\cite{hsiao2024ref}, PVA~\cite{xu2024personalized}, TransRef~\cite{liu2025transref}, OmniGen~\cite{xiao2025omnigen}, and HiFi-Inpaint~\cite{liu2026hifi}.
To evaluate identity preservation, we report Identity Similarity computed via CosFace~\cite{wang2018cosface} alongside perceptual metrics such as FID~\cite{heusel2017gans}, KID~\cite{binkowski2018demystifying, karras2020training}, PSNR, SSIM, and LPIPS~\cite{zhang2018unreasonable}.
For text controllability assessed on whole-face masks using 15 edit prompts~\cite{xu2024personalized}, we measure CLIPScore~\cite{hessel2021clipscore}, ImageReward~\cite{xu2023imagereward}, and Facial Attribute Accuracy(Attr-Acc)~\cite{liu2015deep}.
Note that TransRef is excluded from text-based evaluations due to its lack of text conditioning.

\subsection{Identity-Preserving Face Inpainting}
\subsubsection{Qualitative Results}
\begin{figure}[tb]
  \centering
  \includegraphics[width=0.5\textwidth]{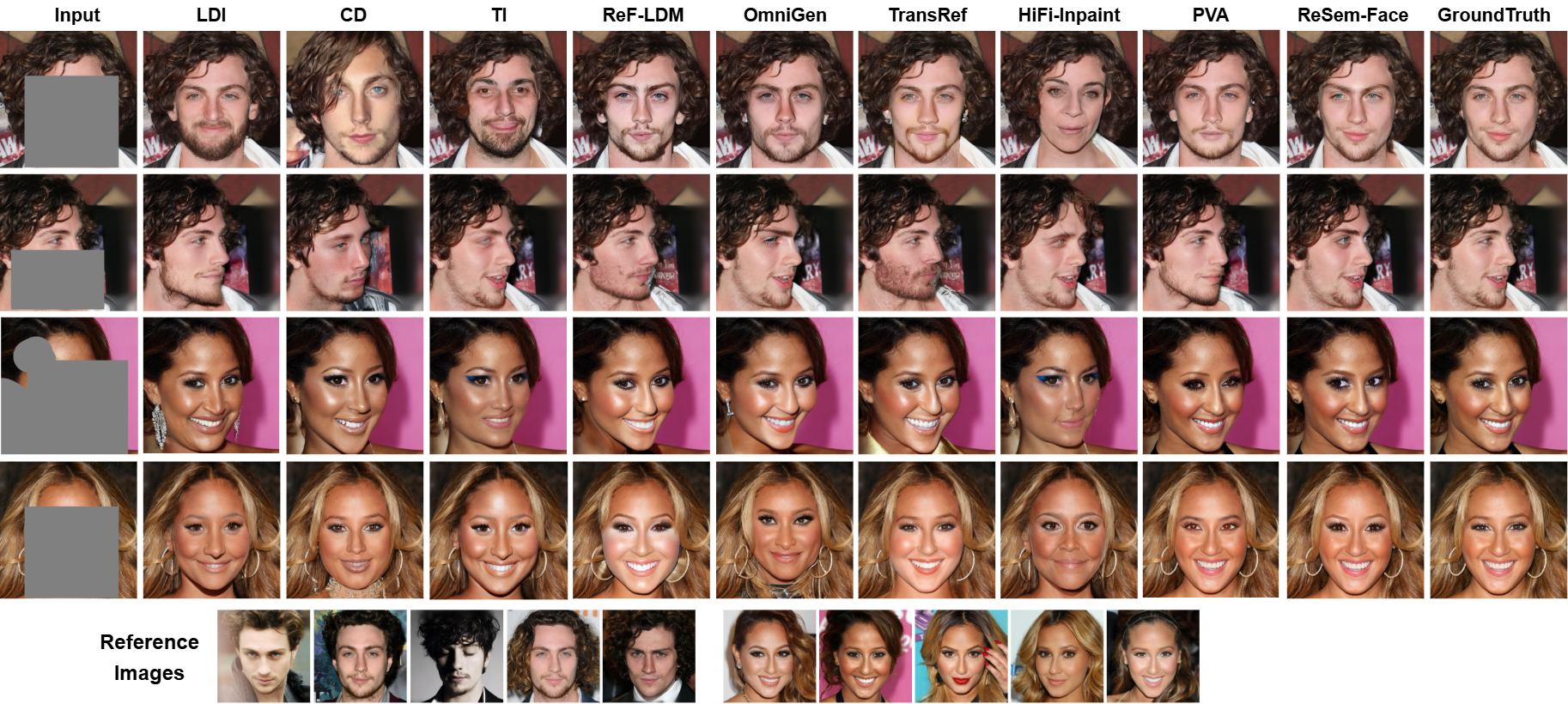}
  \caption{Inpainting results of ReSem-Face and baselines on the test set of CelebAHQ-IDI-5 dataset. 
  }
  \label{fig:baselines}
\end{figure}

Fig.~\ref{fig:baselines} compares ReSem-Face with eight baselines on CelebAHQ-IDI-5.
LDI often suffers from severe identity drift, while Custom Diffusion and PVA improve identity consistency but frequently yield over-smoothed textures or structural inconsistencies under large occlusions.
Although OmniGen and TransRef enhance reference alignment, they still exhibit subtle identity mismatches or boundary artifacts.
In contrast, ReSem-Face generates the most faithful completions across all mask types.
By leveraging the identity-conditioned semantic prior, our method accurately reconstructs identity-defining structures and sharp textures, significantly outperforming baselines in maintaining identity coherence when visual evidence is sparse.

\subsubsection{Quantitative Results}
Tab.~\ref{tab:small} summarizes the quantitative comparisons on CelebAHQ-IDI-5.
First, reference-guided methods substantially outperform generic inpainting and lightweight personalization in identity similarity.
Second, while TransRef achieves high PSNR and SSIM due to its pixel-aligned prompting, ReSem-Face attains the best FID and LPIPS.
This indicates that our method generates more perceptually natural textures compared to pixel-wise optimization.
Most importantly, ReSem-Face achieves the highest ID score of 0.766, surpassing the strongest baseline by a clear margin.
These results confirm that the proposed semantic prior effectively enhances both identity fidelity and perceptual realism under large occlusions.

\begin{table*}[t]
\centering
\resizebox{\textwidth}{!}{
    \setlength{\tabcolsep}{2.5pt} 
    \begin{tabular}{c c c c c c c c c} 
    \toprule
    Method & Venue & FT.Time & ID $\uparrow$ & FID $\downarrow$ & KID $\downarrow$ & PSNR $\uparrow$ & SSIM $\uparrow$ & LPIPS $\downarrow$ \\
    \midrule
    LDI~\cite{rombach2022high}  & CVPR'22   & $-$ & 0.326 & 8.79 & \textbf{2.92} & 24.52 & 0.815 & 0.138 \\
    CD~\cite{kumari2023multi}     & CVPR'23  & $\sim$3min & 0.688 & 15.86 & 7.87 & 26.85 & 0.862 & 0.132 \\
    TI~\cite{gal2023an}       & ICLR'23   & $\sim$6min & 0.615 & 17.50 & 9.12 & 25.40 & 0.830 & 0.138 \\
    ReF-LDM~\cite{hsiao2024ref} & NeurIPS'24 & $-$ & 0.725 & 8.35 & 4.55 & 27.50 & 0.885 & 0.122 \\
    PVA~\cite{xu2024personalized}  & WACV'24  & $\sim$1min & \underline{0.736} & \underline{8.16} & 4.27 & 27.92 & 0.895 & \underline{0.118} \\
    TransRef~\cite{liu2025transref} & Neurocomputing'25 & $-$ & 0.713 & 8.56 & 4.79 & \textbf{28.45} & \textbf{0.910} & 0.128 \\
    OmniGen~\cite{xiao2025omnigen}  & CVPR'25 & $-$ & 0.704 & 9.45 & 4.82 & 26.14 & 0.848 & 0.126 \\
    HiFi-Inpaint~\cite{liu2026hifi}  & CVPR'26 & $-$ & 0.722 & 9.79 & 4.91 & 26.34 & 0.896 & 0.125 \\

    \midrule
    ReSem-Face (Ours) & $-$ & $\sim$1min & \textbf{0.766} & \textbf{7.90} & \underline{3.75} & \underline{28.31} & \underline{0.904} & \textbf{0.116} \\
    \bottomrule
    \end{tabular}
}
\caption{Quantitative results on the test set of CelebAHQ-IDI-5. We quantify the per-identity tuning overhead in the "FT.Time" column using a single RTX A40, where "ID" denotes the identity similarity score. Fine-tuning is limited to 40 steps for all applicable models. KID is scaled by $10^{-3}$. The best and second-best results are highlighted in \textbf{bold} and \underline{underline}, respectively.}
\label{tab:small}
\end{table*}

\begin{table*}[t]
\centering
\resizebox{\textwidth}{!}{
    \setlength{\tabcolsep}{2.5pt}
    \begin{tabular}{ccccccccc}
    \toprule
    Method & Venue & FT.Time & ID $\uparrow$ & FID $\downarrow$ & KID  $\downarrow$ & PSNR $\uparrow$ & SSIM $\uparrow$ & LPIPS $\downarrow$ \\
    \midrule
    LDI~\cite{rombach2022high} & CVPR'22    & $-$ & 0.312 & \underline{17.50} & \textbf{6.85} & 21.45 & 0.715 & 0.172 \\
    CD~\cite{kumari2023multi}   & CVPR'23   & $\sim$3min & 0.585 & 21.30 & 12.40 & 22.10 & 0.742 & 0.165 \\
    TI~\cite{gal2023an} & ICLR'23 & $\sim$6min & 0.525 & 22.80 & 13.50 & 21.50 & 0.725 & 0.175 \\
    ReF-LDM~\cite{hsiao2024ref} & NeurIPS'24 & $-$ & \underline{0.645} & 17.90 & 8.85 & 23.80 & 0.775 & 0.148 \\
    PVA~\cite{xu2024personalized}  & WACV'24  & $\sim$1min & 0.638 & 18.10 & 8.95 & \underline{24.15} & 0.782 & \underline{0.145} \\
    TransRef~\cite{liu2025transref} & Neurocomputing'25 & $-$ & 0.612 & 18.85 & 9.55 & 24.12 & \textbf{0.795} & 0.152 \\
    OmniGen~\cite{xiao2025omnigen} & CVPR'25 & $-$ & 0.605 & 19.42 & 10.12 & 23.05 & 0.768 & 0.158 \\
    HiFi-Inpaint~\cite{liu2026hifi}  & CVPR'26 & $-$ & 0.612 & 17.82 & 8.52 & 23.94 & 0.788 & 0.151 \\
    
    \midrule
    ReSem-Face (Ours)  & $-$ & $\sim$1min & \textbf{0.672}  & \textbf{16.85} & \underline{7.62} & \textbf{24.56} & \underline{0.792} & \textbf{0.142} \\
    \bottomrule
    \end{tabular}
}
\caption{Cross-dataset evaluation on the test set of VGGFace2. The metric definitions and fine-tuning settings are identical to those in Tab.~\ref{tab:small}.}
\label{tab:vggtest}
\end{table*}


\subsubsection{Cross-Dataset Evaluation on VGGFace2}

\begin{figure}[tb]
  \centering
  \includegraphics[width=0.5\textwidth]{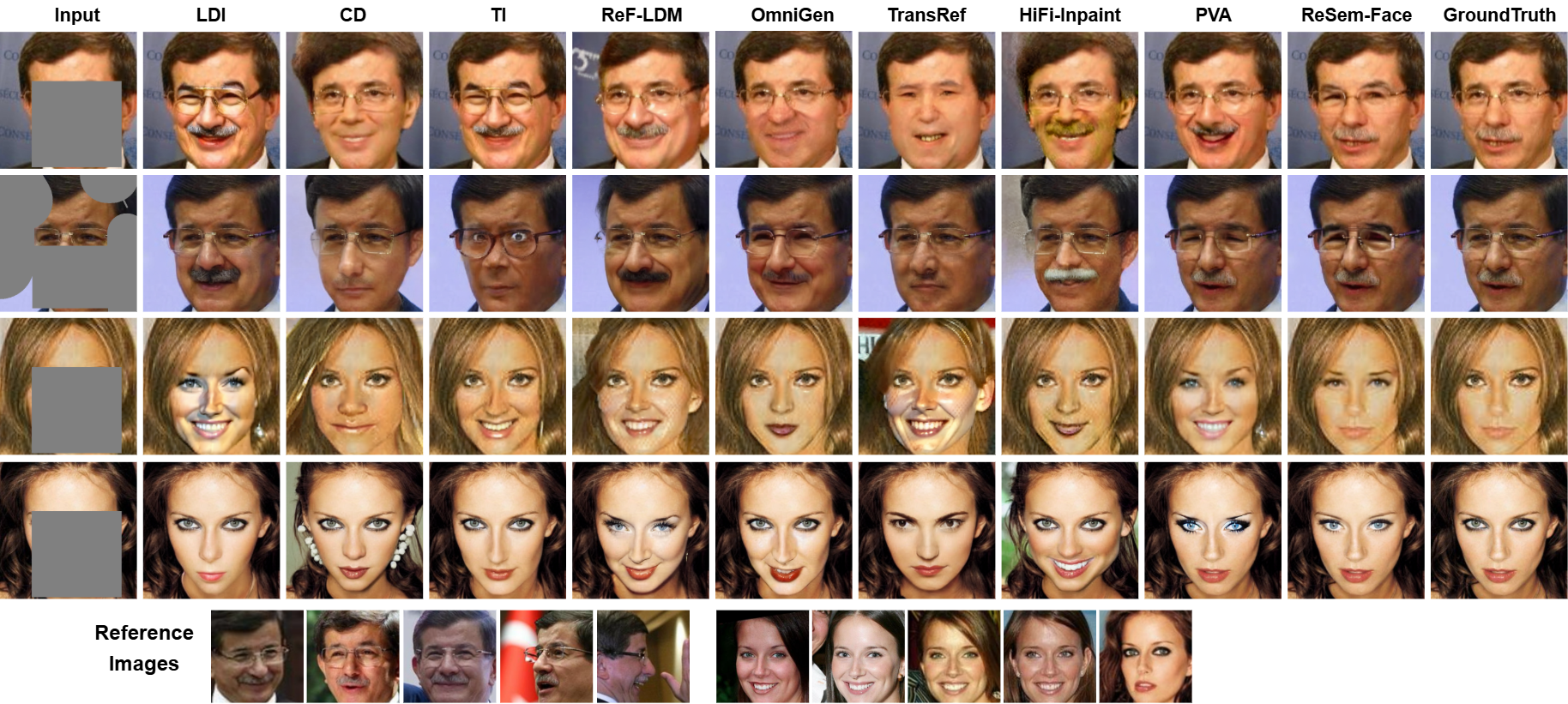}
  \caption{Inpainting results of ReSem-Face and baselines on the test set of VGGFace2 dataset.}
  \label{fig:vggtest_baselines}
\end{figure}

To evaluate generalization capabilities on in-the-wild data, we conduct experiments on the VGGFace2 test set~\cite{cao2018vggface2}, with qualitative comparisons shown in Fig.~\ref{fig:vggtest_baselines}. 
We utilize the official test split with faces aligned and resized to $512 \times 512$, noting that the resulting ground-truth images often exhibit blurriness which affects reference-based metrics.
Quantitative results in Tab.~\ref{tab:vggtest} show that ReSem-Face leads in identity preservation with an ID score of 0.672.
Although all methods suffer from domain gaps and resolution mismatches compared to CelebA-HQ, ReSem-Face consistently achieves the lowest FID and LPIPS.
This confirms that our framework generalizes well to unseen identities and remains robust even when reference images are of lower quality.

\subsubsection{User Study}
Next, we carry out a user study to examine whether the inpainted images conform to human preferences.
In the experiments, we randomly sample 1K images from the test set of CelebAHQ-IDI-5.
We invite 10 evaluators (5 males and 5 females) with diverse education backgrounds: art design (4), psychology (2), computer science (2), and business (2).
We present the evaluators with the masked inputs, reference images, text prompts, and the inpainted results from different methods in a randomized order.
We ask them to assign scores (1$\sim$5) from three aspects:
1) Identity Fidelity: whether the completed face maintains the identity characteristics of the reference images;
2) Text Alignment: whether the generated content aligns with the semantic attributes described in the text prompt;
3) Visual Realism: whether the inpainted region is visually coherent with the surrounding context and structurally natural.
Table~\ref{tab:user_study} summarizes the averaged results of different approaches.
As indicated by the results, ReSem-Face leads the competition by a clear margin against the other baselines in terms of both identity preservation and text-driven editability, aligning best with human perception.

\begin{table}[t]
\centering
\setlength{\tabcolsep}{2.5pt}
\begin{tabular}{cccccc}
\toprule
Method & LDI & CD & OmniGen & PVA & ReSem-Face\\
\midrule
Identity Fidelity & 1.45 & 2.82 & 3.15 & 3.72 & 4.35 \\
Text Alignment    & 2.15 & 2.50 & 3.88 & 3.25 & 4.18 \\
Visual Realism    & 2.90 & 3.20 & 3.65 & 3.58 & 4.40 \\
\bottomrule
\end{tabular}
\caption{User study results on 1K randomly sampled test images from CelebAHQ-IDI-5. We report the average user preference score (1-5), where higher is better.}
\label{tab:user_study}
\end{table}

\subsection{Text Controllability}

\subsubsection{Qualitative Results}

\begin{figure}[t]
  \centering
  \includegraphics[width=0.48\textwidth]{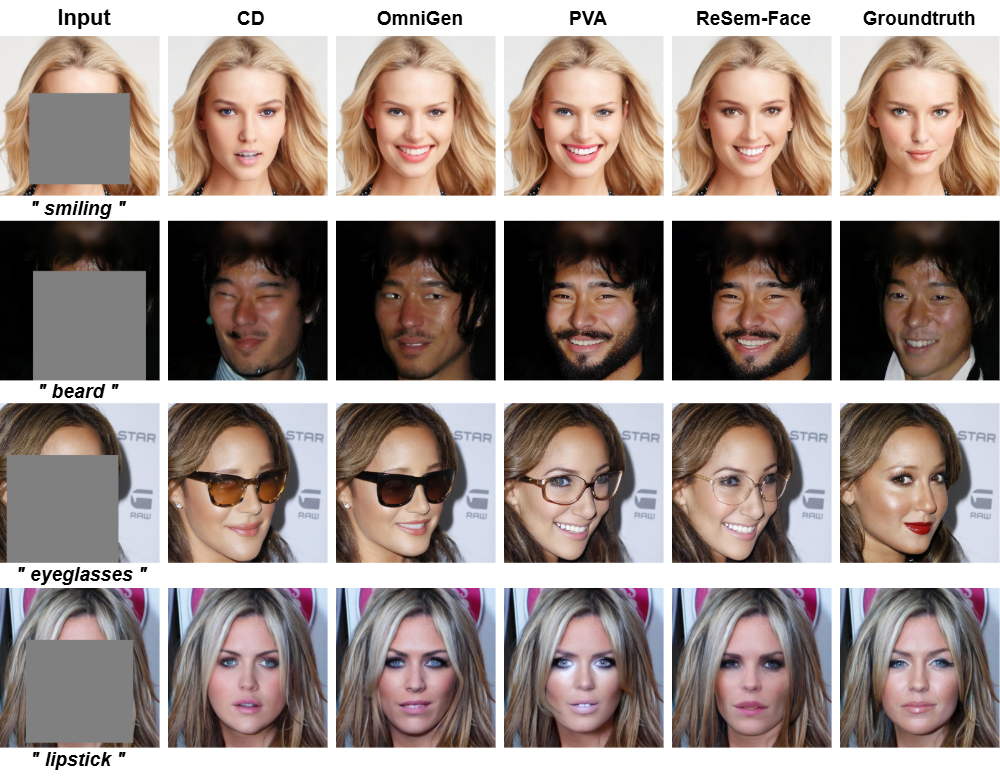}
  \caption{Qualitative comparisons of identity-preserving text-controlled inpainting. Prompts for editing are shown at the bottom of each row. The column tabs, “CD, OmniGen, PVA” denote Custom Diffusion~\cite{kumari2023multi}, OmniGen~\cite{xiao2025omnigen}, PVA~\cite{xu2024personalized}, respectively.
  }
  \label{fig:c-baselines}
\end{figure}

Fig.~\ref{fig:c-baselines} presents qualitative comparisons for text-controlled inpainting. Custom Diffusion responds to prompts but compromises identity under strong edits. PVA maintains stability yet over-smoothes complex attributes. OmniGen captures global structure but struggles to balance fine-grained attributes with identity constraints. In contrast, ReSem-Face delivers the most compelling results across prompts, faithfully executing target attributes while preserving identity-defining structures and natural textures, demonstrating that our semantic prior effectively bridges language control and identity preservation.

\begin{table}[t]
\centering
\setlength{\tabcolsep}{1pt} 
\small
\begin{tabular}{ccccc}
\toprule
Method & ID $\uparrow$ & CLIPScore $\uparrow$ & ImageReward $\uparrow$ & Attr-Acc (\%) $\uparrow$ \\
\midrule
CD & 0.617 & 0.256 & 0.52 & 81.5 \\
OmniGen & 0.602 & 0.289 & \textbf{0.68} & 89.2 \\
PVA & \underline{0.632} & \underline{0.309} & 0.61 & \underline{91.8} \\
\midrule
ReSem-Face & \textbf{0.656} & \textbf{0.318} & \underline{0.65} & \textbf{92.5} \\
\bottomrule
\end{tabular}
\caption{Quantitative results on CelebAHQ-IDI-5 for text-controlled face inpainting. We additionally report ImageReward to evaluate human preference and Attr-Acc for attribute classification accuracy. The best and second-best results are highlighted in \textbf{bold} and \underline{underline}, respectively.}
\label{tab:control}
\end{table}

\begin{table}[t]
\centering
\setlength{\tabcolsep}{1pt} 
\small
\begin{tabular}{ccc ccc cc}
\toprule
\multicolumn{3}{c}{Components} & \multicolumn{3}{c}{Identity-Preserving} & \multicolumn{2}{c}{Text-Controlled} \\
\cmidrule(r){1-3} \cmidrule(lr){4-6} \cmidrule(l){7-8}
\# & SemPrior & ReSemAttn & ID $\uparrow$ & FID $\downarrow$ &  KID $\times 10^{-3}$ $\downarrow$  & ID $\uparrow$ & CLIPScore $\uparrow$ \\
\midrule
1 &         &        & 0.702 & 9.61 & 5.12 & 0.592 & 0.311 \\
2 & \cmark  &        & 0.747 & 9.62 & 5.11 & 0.613 & 0.309 \\
3 & \cmark  & \cmark & \textbf{0.766} & \textbf{7.90} & \textbf{3.75} & \textbf{0.656} & \textbf{0.318} \\
\bottomrule
\end{tabular}
\caption{Component ablation analysis on CelebAHQ-IDI-5. We evaluate the impact of the semantic prior and its injection strategy on both identity-preserving inpainting and text-controlled editing.}
\label{tab:ablation_components}
\end{table}

\subsubsection{Quantitative results}
Tab.~\ref{tab:control} reports quantitative results, including CLIPScore, ImageReward, and Attr-Acc. Baseline methods often sacrifice identity for prompt alignment. In contrast, ReSem-Face achieves the best overall trade-off, obtaining the highest ID similarity, CLIPScore, and Attr-Acc while maintaining competitive ImageReward. These results confirm that our semantic prior enables precise editing without compromising identity consistency.

\subsection{Ablation Study}
\begin{figure}[t]
  \centering
  \begin{subfigure}[b]{0.48\textwidth}
    \centering
    \begin{tikzpicture}
      \begin{axis}[
          width=\linewidth,
          height=5cm,
          xlabel={\# Reference Images},
          ylabel={Identity Similarity ($\uparrow$)},
          xmin=0.8, xmax=6.2,
          ymin=0.55, ymax=0.80, 
          xtick={1, 2, 3, 4, 5, 6},
          yticklabel style={font=\tiny, /pgf/number format/fixed, /pgf/number format/precision=2},
          xticklabel style={font=\tiny},
          xlabel style={font=\footnotesize},
          ylabel style={font=\footnotesize},
          legend style={font=\tiny, at={(0.98,0.02)}, anchor=south east, fill=white, draw=none}, 
          cycle list={} 
      ]
      
      \addplot[only marks, mark=*, mark size=2pt, color=myorange] coordinates {
          (1, 0.685) (2, 0.692) (3, 0.729) (4, 0.753) (5, 0.766) (6, 0.769)
      };
      \addlegendentry{$s=6.0$}
      
      \addplot[only marks, mark=triangle*, mark size=2.5pt, color=myteal] coordinates {
          (1, 0.641) (2, 0.664) (3, 0.696) (4, 0.705) (5, 0.718) (6, 0.722)
      };
      \addlegendentry{$s=3.0$}

      \addplot[only marks, mark=square*, mark size=2pt, color=mydark] coordinates {
          (1, 0.614) (2, 0.621) (3, 0.655) (4, 0.678) (5, 0.681) (6, 0.693)
      };
      \addlegendentry{$s=1.0$}

      \end{axis}
    \end{tikzpicture}
    \caption{Impact of Reference Number}
    \label{fig:sub_ref_num}
  \end{subfigure}
  \hfill 
  \begin{subfigure}[b]{0.48\textwidth}
    \centering
    \begin{tikzpicture}
       \begin{axis}[
          width=\linewidth,
          height=5cm,
          xlabel={Free Guidance Scale $s$},
          xmin=1, xmax=8,
          xtick={1.5, 3.0, 4.0, 5.0, 6.0, 7.5}, 
          xticklabel style={font=\tiny, rotate=0}, 
          ylabel={\textcolor{blue}{Identity Similarity}},
          axis y line*=left,
          ymin=0.55, ymax=0.70, 
          yticklabel style={font=\tiny, /pgf/number format/fixed, /pgf/number format/precision=2},
          xlabel style={font=\footnotesize},
          ylabel style={font=\footnotesize, blue},
          ytick style={blue}
      ]

      \addplot[thick, color=blue, mark=*, mark size=1.5pt]
      coordinates {
          (1.5, 0.682) (3.0, 0.671) (4.0, 0.656) 
          (5.0, 0.605) (6.0, 0.570) (7.5, 0.560)
      };
      \end{axis}

      \begin{axis}[
          width=\linewidth,
          height=5cm,
          xmin=1, xmax=8,
          hide x axis,
          ylabel={\textcolor{red}{CLIPScore}},
          axis y line*=right,
          ymin=0.25, ymax=0.40, 
          yticklabel style={font=\tiny, /pgf/number format/fixed, /pgf/number format/precision=2},
          ylabel style={font=\footnotesize, red},
          ytick style={red}
      ]

      \addplot[thick, color=red, style=dashed, mark=triangle*, mark size=1.5pt]
      coordinates {
          (1.5, 0.275) (3.0, 0.298) (4.0, 0.318) 
          (5.0, 0.338) (6.0, 0.349) (7.5, 0.356)
      };
      \end{axis}
    \end{tikzpicture}
    \caption{ID vs. CLIP Trade-off}
    \label{fig:sub_tradeoff}
  \end{subfigure}
  
  \caption{Ablation studies on the number of reference images and free guidance scale. (a) Increasing the number of reference images steadily improves identity consistency across different free guidance scales. (b) Increasing guidance scale improves text alignment (CLIP) but leads to identity degradation (ID); we choose $s=4.0$ as the optimal trade-off point.}
  \label{fig:ablation_combined}
\end{figure}
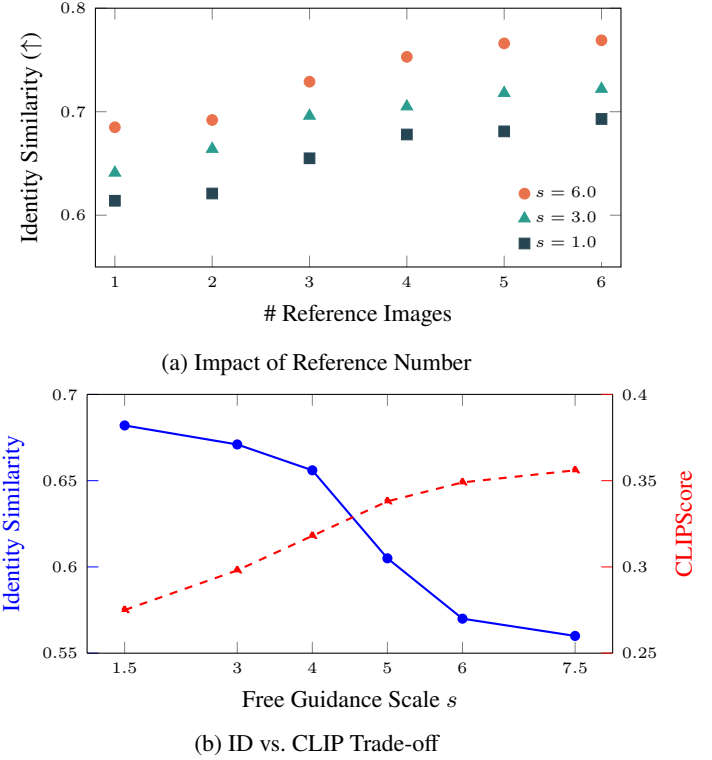

We validate the effectiveness of the proposed semantic prior and its injection strategy through component-wise ablation (Tab.~\ref{tab:ablation_components}) and hyperparameter analysis (Fig.~\ref{fig:ablation_combined}).

\subsubsection{Component Analysis} 
We compare three variants: (1) no semantic prior, (2) semantic prior without injection using auxiliary supervision only, and (3) the full model with ReSemAttn injection. 
Results in Tab.~\ref{tab:ablation_components} show that removing the semantic prior entirely leads to a significant performance drop, where the ID score decreases from 0.766 to 0.702, confirming that high-level semantics are essential when visual evidence is missing. 
Furthermore, Var.~2 yields only marginal gains, verifying that explicit injection via ReSemAttn is crucial for stabilizing identity reconstruction. 
Notably, our full model (Var.~3) utilizes a decoupled semantic architecture consisting of $\mathcal{H}_{geo}$ and $\mathcal{H}_{tex}$ to mitigate ``semantic blurring'' where identity-rich features interfere with rigid geometric parsing. 
This architectural choice ensures stable semantic injection and structural consistency without compromising the fidelity of identity texture, leading to the superior performance observed in ReSem-Face.

\subsubsection{Hyperparameter Analysis} 
We further investigate the impact of key hyperparameters in Fig.~\ref{fig:ablation_combined}. 
First, regarding the number of reference images (Fig.~\ref{fig:ablation_combined}(a)), we observe that increasing the references from 1 to 6 consistently boosts identity similarity. Notably, using multi-reference guidance maintains high identity fidelity even under strong guidance scales.
Second, regarding the conflict between text and identity (Fig.~\ref{fig:ablation_combined}(b)), we analyze the trade-off under varying guidance scales ($s$). While higher scales improve CLIPScore, they often degrade identity preservation. 
Our analysis identifies $s=4.0$ as the optimal sweet spot, achieving precise attribute editing with minimal loss in identity fidelity.


\section{Conclusion}
In this paper, we presented ReSem-Face, a semantic-enhanced diffusion framework for identity-preserving and text-controllable face inpainting under large occlusions. By combining identity-aware semantic pre-inpainting, Reference Identity Attention, and Reference Semantic Attention, ReSem-Face provides complementary identity and semantic constraints during denoising. Experiments on CelebAHQ-IDI-5 and VGGFace2 demonstrate its superiority in identity fidelity, visual realism, and text controllability over representative baselines.


\bibliography{aaai2027}


\end{document}